\documentclass[runningheads]{llncs}
\usepackage[dvipdfmx]{graphicx}
\usepackage{color}

\newcommand{\Tag}[1]{{\it #1}}

\usepackage{multicol}
\usepackage{multirow}
\usepackage[subrefformat=parens]{subcaption}
\usepackage{array} 
\usepackage{here}

\begin{document}
\title{Impression-CLIP: Contrastive Shape-Impression Embedding for Fonts}
%
%
\author{Yugo Kubota\and
Daichi Haraguchi\orcidID{0000-0002-3109-9053}\and\\
Seiichi Uchida\orcidID{0000-0001-8592-7566}}
\authorrunning{Y. Kubota et al.}

\institute{Kyushu University, Fukuoka, Japan
\email{yugo.kubota@human.ait.kyushu-u.ac.jp}}
\maketitle              
\begin{abstract}
Fonts convey different impressions to readers.
These impressions often come from the font shapes.
However, the correlation between fonts and their impression is weak and unstable because impressions are subjective.
To capture such weak and unstable cross-modal correlation between font shapes and their impressions, we propose Impression-CLIP, which is a novel machine-learning model based on CLIP (Contrastive Language-Image Pre-training).
By using the CLIP-based model, font image features and their impression features are pulled closer, and font image features and unrelated impression features are pushed apart.
This procedure realizes co-embedding between font image and their impressions.
In our experiment, we perform cross-modal retrieval between fonts and impressions through co-embedding. The results indicate that Impression-CLIP achieves better retrieval accuracy than the state-of-the-art method.
Additionally, our model shows the robustness to noise and missing tags.

\end{abstract}







\section{Introduction}


\begin{figure}[t] 
\includegraphics[width=\textwidth]{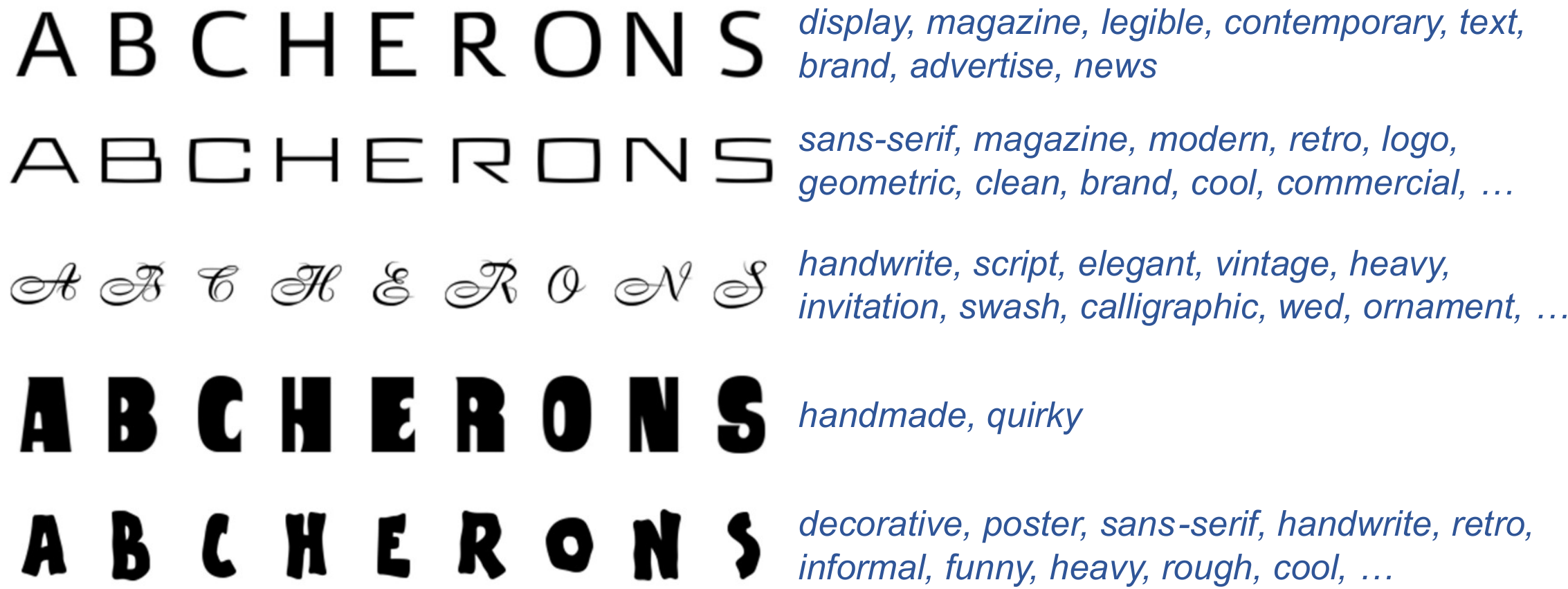}
\caption{Font styles and their impression tags.}
\label{fig:font-example}
\end{figure} 

The shape of machine-printed letters varies depending on fonts. Even for the same letter `A,' there are huge shape variations. For example, when printing `A' using a serif font, its diagonal strokes often end with a short decorative line called a serif. Nowadays, such huge shape variations are available as font collections. Typographic experts (or even non-experts) carefully choose appropriate fonts for their products.\par
%

Different fonts give different impressions to readers. Fig.~\ref{fig:font-example} shows several font examples and their impression tags at \url{MyFonts.com}. Interestingly, these simple shapes (i.e., stroke contours) with neither photo-realistic nor color elements can give various impressions. Moreover, we get such impressions from detailed shape structures independent of the whole shape of the letter itself (e.g., `A'). Namely, an impression from the letter `A' does not come from its basic structure with two diagonal strokes and a horizontal stroke but from finer structures, such as local contour curvature, stroke thickness, corner shapes, serif, etc.\par

%
Dependency between font shapes and their impressions will not be random; in other words, they have a mutual correlation. Understanding the correlation will be beneficial not only from an application viewpoint but also from a scientific viewpoint. In fact, the correlation will be useful for many applications, such as automatic typographic design, impression-based font retrieval, impression-based font generation, etc. The correlation will be useful to understand our cognitive mechanism --- like the well-known bouba/kiki effect\footnote{\url{ https://en.wikipedia.org/wiki/Bouba/kiki\_effect}}, it will reveal our cognitive characteristics on shape perception.\par

However, the correlation analysis between font shapes and impressions is not straightforward for the following issues.
\begin{itemize}
    \item Impressions are subjective. The impression \Tag{heavy} is expected to correlate strongly with thick strokes for most people. In contrast, the impression \Tag{cool} will correlate with different shapes depending on subjective thoughts and personal experiences. A font with very thin strokes can be  \Tag{cool} for several persons, and a font with very thick strokes can also be \Tag{cool} for others.  
    \item Impressions are often noisy. We will use a font-impression dataset collected from \url{MyFonts.com}. In the dataset, impression tags are attached to individual fonts. 
    The fonts and impressions in Fig.~\ref{fig:font-example} are examples. These tags are attached by crowd-sourcing; anonymous people (including non-experts) freely attach them according to their thoughts and intentions. 
    \item There are many missing tags. Even if a font has very heavy strokes, it will not be tagged as \Tag{heavy} unless someone voluntarily tags it. 
\end{itemize} 
Unfortunately, such instability is inevitable when we deal with impressions. If one attempts to develop a new font dataset tagged with impressions under the careful guidance of font experts, the second and third issues could be relaxed. (AMT dataset~\cite{chen2019large} used in this paper is an example of such a more elaborated dataset.) However, as the first issue indicates, impressions are inherently subjective, and there can be no unanimous ground-truth. \par
%

This paper proposes a novel machine-learning model, {\em Impression-CLIP}, for capturing such a weak and unstable cross-modal correlation between font shapes and impressions.
Impression-CLIP is based on CLIP (Contrastive Language-Image Pre-training)~\cite{radford2021learning}, which is a well-known cross-modal embedding model based on a neural network. By CLIP, images and texts share the same feature space. More importantly, by using {\em contrastive learning}, each ``positive'' image-text pair showing a common situation becomes closer in the feature space, whereas each ``negative'' pair showing different situations becomes further. For example, an image capturing birds in the sky and the text ``birds in the sky'' will be embedded as similar feature vectors. Consequently, CLIP can emphasize the positive and negative correlations between image and text modalities. 
\par


\begin{figure}[t] 
\includegraphics[width=\textwidth]{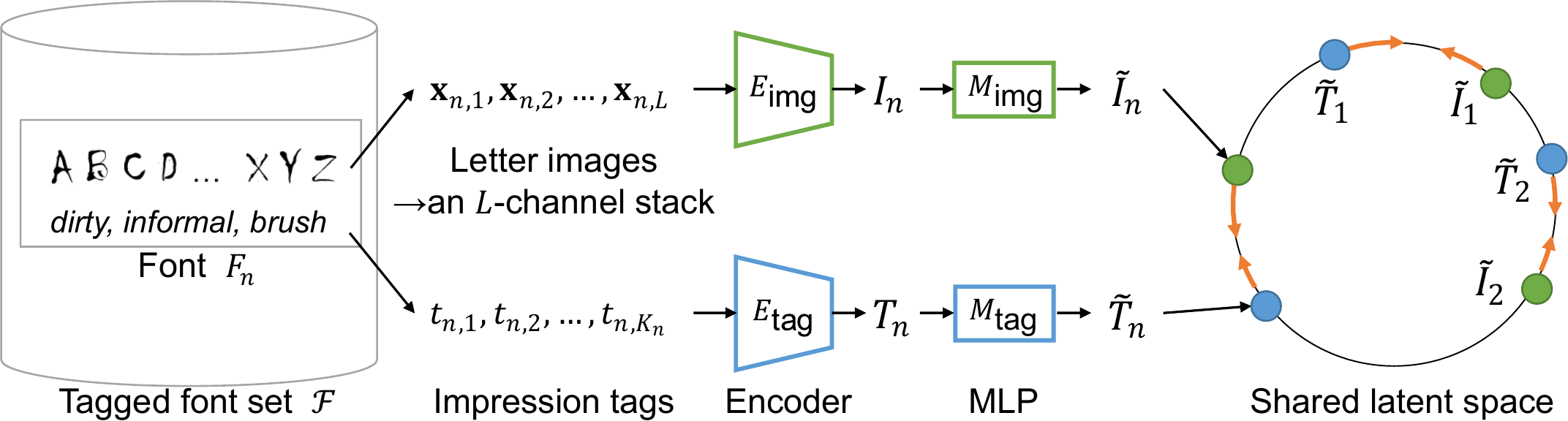}
\caption{Overview of Impression-CLIP, contrastive shape-impression embedding.} \label{fig:overview}
\end{figure} 

Our Impression-CLIP proves that the CLIP framework is also useful for capturing the correlation between font shapes and impressions. Image and text modalities in CLIP correspond to font shape and impression modality in Impression-CLIP, respectively.
Fig.~\ref{fig:overview} gives an overview of Impression-CLIP. (Here, the feature space shared by the font shapes and impressions is depicted as a circle since all feature vectors are normalized to have a norm 1. In reality, the feature space forms a hyper-sphere.) Like the successful applications of CLIP, we can expect that Impression-CLIP will capture weak and unstable correlations between shapes and impressions by training its model with many shape-impression pairs in a contrastive manner. \par
%
%
Through multiple experiments with about 20,000 different fonts, we show that Impression-CLIP can enhance the positive and negative correlations between font shapes and impressions, as expected. We confirmed that impression-based font retrieval is possible by a simple nearest-neighbor search in the CLIP feature space. We compared our results with the state-of-the-art method, which we call Cross-AE~\cite{kang2022shared} in this paper.  The results show that Impression-CLIP outperforms Cross-AE in qualitative and quantitative evaluations. \par

The main contributions of this paper are summarized as follows:
\begin{itemize}
    \item We propose Impression-CLIP to understand the cross-modal correlation between shapes and impressions of individual fonts.
    \item Through experimental validations, we confirmed that the positive and negative correlations between shapes and impressions are enhanced successfully from their original correlations. 
    \item As a practical application example of Impression-CLIP, we performed 
    impression\allowbreak-based font retrieval experiments and found that ours achieved better retrieval accuracy than the state-of-the-art method. 
\end{itemize}

\section{Related Work}
\subsection{Font Impression} 
Many researchers have been interested in the cross-modal correlation between font shapes and impressions. Even from the early 1920s, psychologists started subjective experiments to reveal the correlation~\cite{davis1933determinants,poffenberger1923study}. Research has also been conducted from more practical aspects; for example, Henderson~\textit{et al.}~\cite{henderson2004impression} analyzed the correlation for a marketing purpose, and, more recently, Choi ~\textit{et al.}~\cite{choi2019emotype} did it for emotion expression in SNS messages.
\par


Various application tasks, such as font retrieval and font generation, are conducted to leverage cross-modal correlations between font shapes and impressions.
O'Donovan~\textit{et al.}~\cite{o2014exploratory} collected the impression tags of fonts through crowd-sourcing and proposed an interface to retrieve font from the impression.
Chen~\textit{et al.}~\cite{o2014exploratory} created a large-scale font dataset collected from \url{MyFonts.com}; this dataset is much larger than \cite{o2014exploratory} and includes
impression tags attached by anonymous users of MyFonts. Ueda~\textit{et al.}~\cite{ueda2021parts,ueda2022font} utilized this MyFonts dataset for impression estimation of fonts. Font impressions have also been utilized for several impression-based font generation tasks~\cite{matsuda2021impressions2font,matsuda2022font,wang2020attribute2font}.
\par

\subsection{CLIP and Its Application}
CLIP~\cite{radford2021learning} is a cross-modal embedding (i.e., co-embedding) model based on neural networks. CLIP is based on contrastive learning. In the general contrastive learning framework, we assume positive pairs and negative pairs. Each positive pair consists of two data we want to make similar, and each negative pair consists of two data we do not want to. After training the model to make positive pairs more similar and negative pairs more dissimilar, the model can represent data so that its distribution satisfies our 
positive/negative requests. In CLIP, pairs are multimodal. For example, each positive pair consists of an image $\mathbf{x}$ and the caption $\mathbf{t}$ of the image, whereas each negative pair consists of an image $\mathbf{x}$ and the caption $\mathbf{t}'$ of a different image $\mathbf{x}'$. After its contrastive learning, $\mathbf{x}\sim \mathbf{t}$ and $\mathbf{x'}\sim \mathbf{t'}$, whereas $\mathbf{x}\neq \mathbf{t'}$ and $\mathbf{x'}\neq\mathbf{t}$ in their representations.
\par

CLIP has been used for various tasks in the so-called ``vision and language'' research field using its high cross-modal embedding ability. 
For example, text-to-image generation~\cite{frans2022clipdraw,izumi2022zero,Kwon_2022_CVPR} has been conducted by leveraging the cross-modal embedding as $\mathbf{t}\to \mathbf{x}$.
Furthermore, the consistent relation $\mathbf{t}\leftrightarrow \mathbf{x}$ can be utilized for text-based image retrieval, such as text-video retrieval~\cite{luo2022clip4clip,ma2022x} and artwork retrieval~\cite{Conde_2021_CVPR}. Text features by CLIP have been used for prompts of diffusion models~\cite{Rombach_2022_CVPR} because CLIP can embed a phrase (with multiple words) into a single vector while representing the meaning of the phrase. 



\subsection{Co-embedding Shape and Impression of Fonts}\label{sec:related-co-emb} 
A few studies attempted to realize co-embedding shape and impression of fonts and used the co-embedding results for font retrieval. Kulahcioglu~\textit{et al.}\cite{kulahcioglu2020fonts} and Choi~\textit{et al.}\cite{choi2019assist} realized the co-embedding for conducting both impression-based font retrieval and impression estimation of fonts. The former created an embedding matrix of font-impression relationship, whereas the latter used a multimodal deep Boltzmann machine for co-embedding. \par
%
More recently, Kang~\textit{et al.}~\cite{kang2022shared} proposed a co-embedding method, which will be called Cross-AE in this paper. Cross-AE uses two autoencoders, one for font shape (i.e., font image) and the other for impression tags. These auto-encoders are 
co-trained to have similar latent representations for the shape and impression of the same 
font.\par

Inspired by \cite{kang2022shared}, we utilize CLIP to co-embed
font shapes and impressions into the same latent feature space. The most important difference between our Impression-CLIP and Cross-AE~\cite{kang2022shared} is that the former embeds two modalities into the same feature space more contrastive with the help of the CLIP framework. 
Specifically, Cross-AE is trained by considering only positive pairs (where each pair is a set of character images and a set of impression tags from the same font). In contrast, Impression-CLIP also considers negative pairs (where each pair comes from different fonts).
This difference in the training scheme gives a significant difference between those methods; in fact, experimental results show a large superiority of Impression-CLIP over Cross-AE.

\section{Impression-CLIP}

\subsection{Overview\label{sec:overview-of-Impression-CLIP}}  
Fig.~\ref{fig:overview} shows an overview of Impression-CLIP. 
Assume we have a set of fonts $\mathcal{F}=\{F_n\}$ and each font
comprises a set of $L$ letter images and a set of $K_n$ impression tags, that is, $F_n = \{\mathbf{x}_{n,1}, \ldots, \mathbf{x}_{n,L}, t_{n,1},\ldots, t_{n,K_n}\}$. Since we deal with the capital letters of the Latin alphabet (`A'-`Z') in this paper, $L=26$.
The $L$ letter images are treated as an $L$-channel stack $X$ and converted into a feature vector $I_n$ by an encoder $E_\mathrm{img}$. The $K_n$ impression tags are converted as a feature vector $T_n$ by another encoder $E_\mathrm{tag}$. The features $I_n$ and $T_n$ are further converted into latent feature vectors $\tilde{I}_n$ and  $\tilde{T}_n$ by different multi-layer perceptrons (MLPs) $M_\mathrm{img}$ and $M_\mathrm{tag}$, respectively. We assume all latent feature vectors are $d$-dimensional and normalized so that their norm is 1, that is, $\|\tilde{I}_n\|=\|\tilde{T}_n\|=1$. Consequently, the font image features $\{\tilde{I}_n\}$ and the font impression features $\{\tilde{T}_n\}$  share the latent space and distribute in the $d$-dimensional hyper-sphere. 
\par
%
As noted above, the Impression-CLIP model is trained in a contrastive learning framework, whose goal is to make $\tilde{I}_n\sim\tilde{T}_n$ for each font $F_n$ and $\tilde{I}_n\not\sim\tilde{T}_m$ for different fonts $F_n$ and $F_m$ ($n\neq m)$. The trained Impression-CLIP will inherit various merits of CLIP. Since latent vectors in the different modalities share the same $d$-dimensional feature space while keeping their cross-modal similarity and dissimilarity, it is possible to realize impression-based font retrieval by simple nearest-neighbor search in the $d$-dimensional space. The nearest-neighbor search also enables impression estimation for an unknown font (such as a font generated by some generative AI models). In addition to these practical applications, we can realize a more reliable cross-modal correlation analysis between shapes and impressions because the correlation is enhanced by contrastive learning in Impression-CLIP.
\par

\subsection{Extracting Font Shape Features}  

The image encoder $E_\mathrm{img}$ is pre-trained in the standard autoencoder framework. 
Specifically, the autoencoder is trained to reconstruct an $L$-channel image input $X$ via a $d$-dimensional latent vector $I_n$. Again, an $L$-channel image $X$ is the stack of all $L$ letter images from a certain font. This means that we define a font shape feature common for all alphabet letters. This is important for the following reasons. First, the impressions are attached not to individual letters but to all letters. Second, font designers create a font
by considering the style consistency over all letters.
Third, if we treat only each letter independently, the effect of the letter shape becomes more dominant than the font style. (For example, the letter `O' never has serifs, even if it comes from a serif font.)
\par

\subsection{Extracting Font Impression Features}\label{sec:pretraining-font-imprssion-feature}
We employ a text encoder by the original CLIP~\cite{radford2021learning} as $E_\mathrm{tag}$. We use its pretrained version without fine-tuning because it is already (pre)trained with a huge corpus. A preliminary experiment shows that even special terms in typography (such as \Tag{serif}) are appropriately converted into feature vectors. \par
For feeding the $K_n$ impression tags $\{t_{n,1},\ldots, t_{n,K_n}\}$ into the text encoder, we need to convert them as a single prompt. More specifically, we use the prompt:
 ``\texttt{First, second, third, $\ldots$, and tenth impressions are $t'_{n,1}$, $t'_{n,2}$, $t'_{n,3}$, $\ldots$, and $t'_{n,10}$, respectively.}''  Here, $t'_{n,1},\ldots,t'_{n,K_n}$ is the reordered version of $t_{n,1},\allowbreak \ldots, \allowbreak t_{n,K_n}$ according to the frequency of the tags in the whole dataset $\mathcal{F}$. Therefore, $t'_{n,1}$ is the most frequent tag among $\{t_{n,1},\allowbreak \ldots, \allowbreak t_{n,K_n}\}$ and  $t'_{n,K_n}$ is the least. As indicated by this prompt, we used the ten most frequent tags even when $K_n>10$; this approach ignores less frequent tags by assuming they are noisy tags.\footnote{Assume the case with three tags instead of ten tags. Also, assume that the $n$th font has $\{t_{n,1},\ldots, t_{n,4}\}=\{\Tag{hand}, \Tag{cute}, \Tag{kid}, \Tag{informal}\}$ and each tag has 797, 422, 356, and 2356 occurrences, respectively.
Then, $\{t'_{n,1}, t'_{n,2}, t'_{n,3}\}=
\{\Tag{informal}, \Tag{hand}, \Tag{cute}\}$ and the prompt becomes ``{\tt First, second, and third impressions are informal, hand, and cute, respectively.''}}
If $K_n<10$, we use a shorter prompt containing all $K_n$ tags.  
\par




\subsection{Contrastive Learning for Co-embedding Styles and Impressions}  
As noted section~\ref{sec:overview-of-Impression-CLIP}, Impression-CLIP performs contrastive learning to make $\tilde{I}_n\sim\tilde{T}_n$ for each font $F_n$ and $\tilde{I}_n\not\sim\tilde{T}_m$ for $F_n$ and $F_m$ ($n\neq m)$.
In more detail, $M_{\mathrm img}$ and  $M_{\mathrm tag}$ are trained so that the latent representations $\tilde{I}_n$ and $\tilde{T}_n$ are pulled closer gradually, and those of $\tilde{I}_n$ and $\tilde{T}_m$ are pushed away. Note that the encoders $E_\mathrm{img}$ and $E_\mathrm{tag}$ are frozen during contrastive learning of $M_{\mathrm img}$ and  $M_{\mathrm tag}$.
Both font image and impression feature vectors are normalized, that is, $\|\tilde{I}_n\|=\|\tilde{T}_n\|=1$.  Then, the similarity between the feature vectors is defined as $S_{i,j} = \tilde{I}_i\cdot\tilde{T}_j$, which is equivalent to the cosine similarity. Consequently, in the ideal case, $S_{n,n} = 1$ and $S_{n,m} = -1$ $(n\neq m)$ are required for associating a font image and its impression feature vectors.


We employ a symmetric cross-entropy (SCE)~\cite{radford2021learning} for contrastive learning. SCE loss can be seen as a cross-entropy loss for an $B$-class classification problem, where $B$ is a batch size in training. Given an image feature $\tilde{I}_n$, choosing $\tilde{T}_n$ among $B$ impression features in the batch is treated as the correct classification result. Similarly, given an impression feature  $\tilde{T}_n$, choosing $\tilde{I}_n$ among $B$ image features is correct. Based on this strategy, SCE loss is defined as below. 
\begin{eqnarray*}
L_{\mathrm{SCE}} &=& (L_{\mathrm{img}}+L_{\mathrm{tag}})/2,\\
L_{\mathrm{img}} &=& -\sum_{n}^{B}\log\frac{\exp(\overline{S}_{n,n})}{\sum_{m}^{B}\exp(\overline{S}_{n,m})},\\
L_{\mathrm{tag}} &=& -\sum_{n}^{B}\log\frac{\exp(\overline{S}_{n,n})}{\sum_{m}^{B}\exp(\overline{S}_{m,n})},
\end{eqnarray*}
where $L_{\mathrm{img}}$ is a loss function for choosing the correct $n$th impression from the batch $\{ 1,\ldots,n,\ldots,B\}$ for a given $n$th image, whereas $L_{\mathrm{tag}}$ is vice versa. This loss $L_{\mathrm{SCE}}$
reflects the above strategy because it is minimized by maximizing ${S}_{n,n}$ and minimizing ${S}_{n,m} (n\neq m)$.
The modified similarity $\overline{S}_{i,j}$ is calculated as $\overline{S}_{i,j} = S_{i,j}\cdot\exp(\tau)$, where $\tau$ is a trainable hyperparameter called temperature and useful for efficient training. 

\section{Experimental Settings}
\subsection{Dataset~\cite{chen2019large}}\label{sec:dataset}


We use the MyFonts dataset by Chen~\textit{et al.}~\cite{chen2019large}, which consists of font images and their impression tags collected from \url{Myfonts.com}. Fig.~\ref{fig:font-example} shows examples of font images and their impression tags. The dataset originally consists of 18,815 fonts.
Since we use $L=26$ capital letters, we exclude small-capital fonts and dingbat fonts. We also exclude fonts with no tags. As a result, we used 17,154 fonts in our experiment, and they are split into 13,733 training fonts, 1,711 validation fonts, and 1,710 testing fonts. 
%
In the dataset, 1,824 different impression tags. As a preprocessing for impression tags, we eliminate noisy tags those not included in the Google News dataset. Then, by the prompt generation rule of Section~\ref{sec:pretraining-font-imprssion-feature}, we further keep only ten frequent tags (and eliminate the other tags) for the fonts with more than 10 tags.
Finally, we eliminate the tags with less than 50 occurrences in the dataset. After the eliminations, we used 223 impression tags in the following experiment. Each font has 1 to 10 impression tags; the average and median numbers are 7.1 and 8, respectively.
\par




We additionally used the AMT dataset, which is a subset of the MyFonts dataset. 
In AMT, three fonts with the same impression tag (say \Tag{elegant}) in their tag lists are randomly selected first to form a group. Then, three crowd-workers select the best font among the three fonts for the impression. If they unanimously select the same font for the tag, this group is kept, and the selected font is considered to have a strong correlation to the tag; otherwise, the group is discarded. After multiple group formulation attempts, AMT finally contains 1,041 groups. Consequently, each group contains one font strongly correlated to a tag and two fonts weakly correlated to the tag.\par
%

\subsection{Implementation Details}  
In pre-training the autoencoder for obtaining image encoder $E_\mathrm{img}$, we used $L_1$ loss between the input image $X$ and the reconstructed one $\hat{X}$, that is $L_1=\|X-\hat{X}\|_{1}$. The encoder consists of nine convolution layers with residual connection, and the decoder consists of three deconvolution layers. We introduced fully connected layers in the middle layer between the encoder and decoder to set the dimension $d$ of image features to $512$, which is the same dimension as the impression features. The size of the input image is $64\times64$ with $L=26$ channels.
In pre-training the image autoencoder and training Impression-CLIP, we used the Adam optimizer with a learning rate of $10^{-4}$. The batch sizes in each pre-training and training are set to 16 and 8,192, respectively.
\par




\subsection{Comparative Method and Evaluation Metrics\label{sec:metrics}}
We compared our Impression-CLIP with Cross-AE~\cite{kang2022shared}.
As noted in Section~\ref{sec:related-co-emb}, the main difference between ours and Cross-AE is that the former is truly contrastive with positive and negative pairs, whereas the latter does not care about negative pairs. In other words, the latter has no function to make $\tilde{I}_n\not\sim\tilde{T}_m$ for $F_n$ and $F_m$ ($n\neq m)$. Another difference is 
the structure of text encoders; Impression-CLIP uses the text encoder of CLIP, whereas Cross-AE uses its original text encoder with word2vec and the Deep Set model~\cite{kang2022shared}. 
\par
One might think that we need to re-implement Cross-AE with the recent CLIP text encoder for a fair comparison. However, it is impossible. Cross-AE relies on auto-encoders, and to use the CLIP text encoder in the Cross-AE framework, we need to have a CLIP text ``decoder'' that reconstructs original texts from the encoded latent vector $\tilde{T}_n$. However, no CLIP text decoder is available in the current CLIP framework, so we follow the original implementation of Cross-AE~\cite{kang2022shared}. Note that both models use the same encoder architecture (and the same $L$-channel input) as $E_\mathrm{img}$.
\par

In the later experiment, we evaluated cross-modal retrieval tasks: impression-based font image retrieval and image-based impression retrieval (i.e., estimation). As evaluation metrics of retrieval results, we use mean average precision (mAP) and average retrieval rank (ARR). mAP $\in (0,1]$ is the most common metric of information retrieval. ARR evaluates the performance by a retrieval experiment based on a nearest-neighbor search within the test dataset. 
Specifically, for ARR, we first derive the latent vectors $\tilde{I}_n$ and $\tilde{T}_n$ for each test font $F_n,\ n=1, \ldots, 1710$. Then, for each $\tilde{I}_n$, we determine the similarity rank of $\tilde{T}_n$ among 1710 vectors $\{\tilde{T}_k, k=1,\ldots, 1710\}$. (Here, the inner-product $\tilde{I}_n\cdot\tilde{T}_k$ is used as the similarity.)  In the ideal case, $\tilde{T}_n$ becomes the most similar text vector, and we have the rank as 1. 
In the worst case, $\tilde{T}_n$ becomes the least similar text vector, and we have the rank as 1710. ARR (img$\to$ tag) $\in [1, 1710]$ is given by averaging these ranks for all $\tilde{I}_n$. We also have ARR (tag$\to$ img) $\in [1, 1710]$ by interchanging $\tilde{I}_n$ and $\tilde{T}_n$.
\par 

\section{Experimental Results}
\subsection{Visualization of Feature Distributions}
\begin{figure}[t]
    \begin{tabular}{cc}
        \begin{minipage}[t]{0.47\textwidth}
            \centering
            \includegraphics[width=\linewidth]{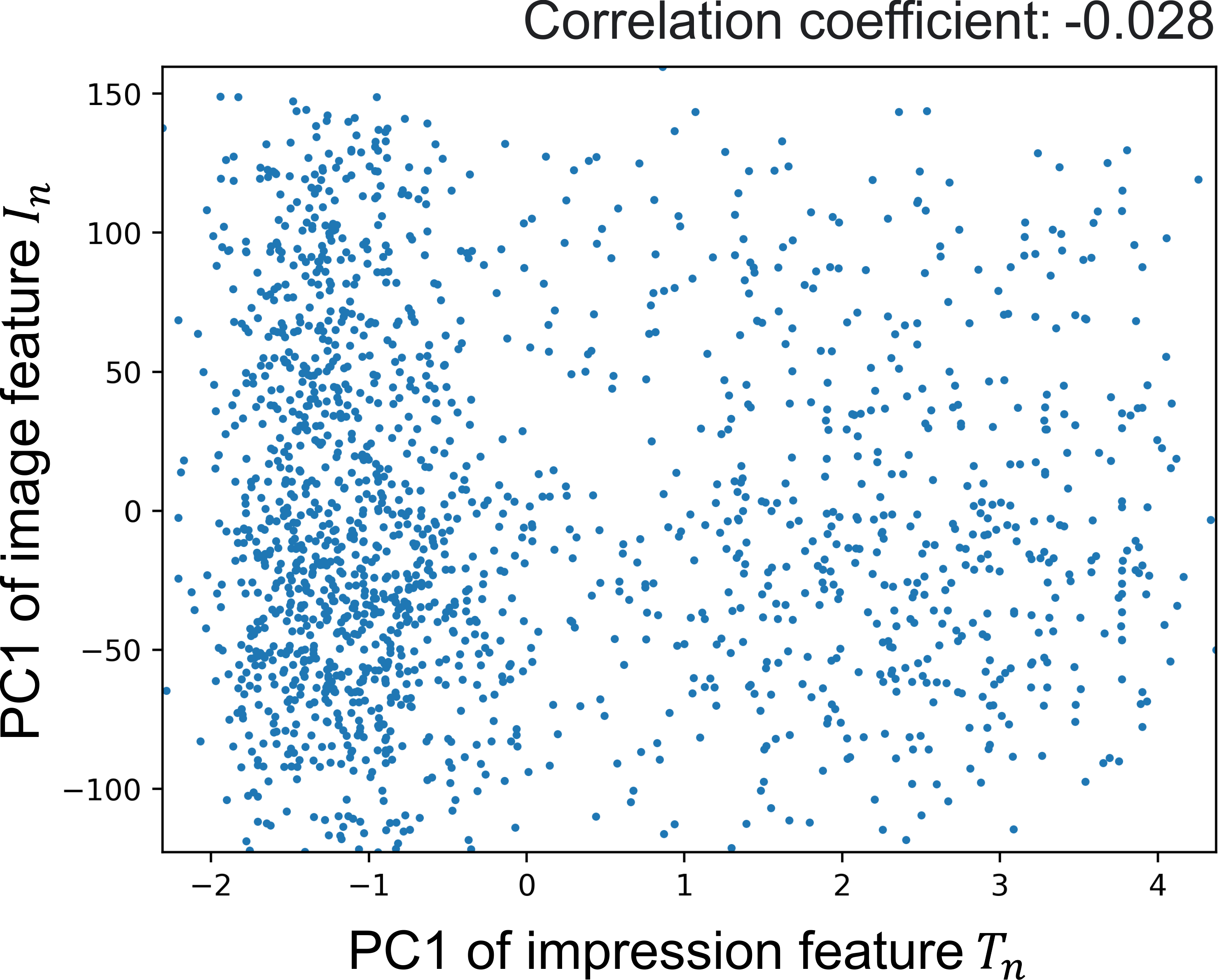} \\[-1mm]
            \subcaption{Before contrastive learning}
            \label{imp-img_CLIP_before}
        \end{minipage} &
        \hspace{0.035\columnwidth}
        \begin{minipage}[t]{0.47\textwidth}
            \centering
            \includegraphics[width=\linewidth]{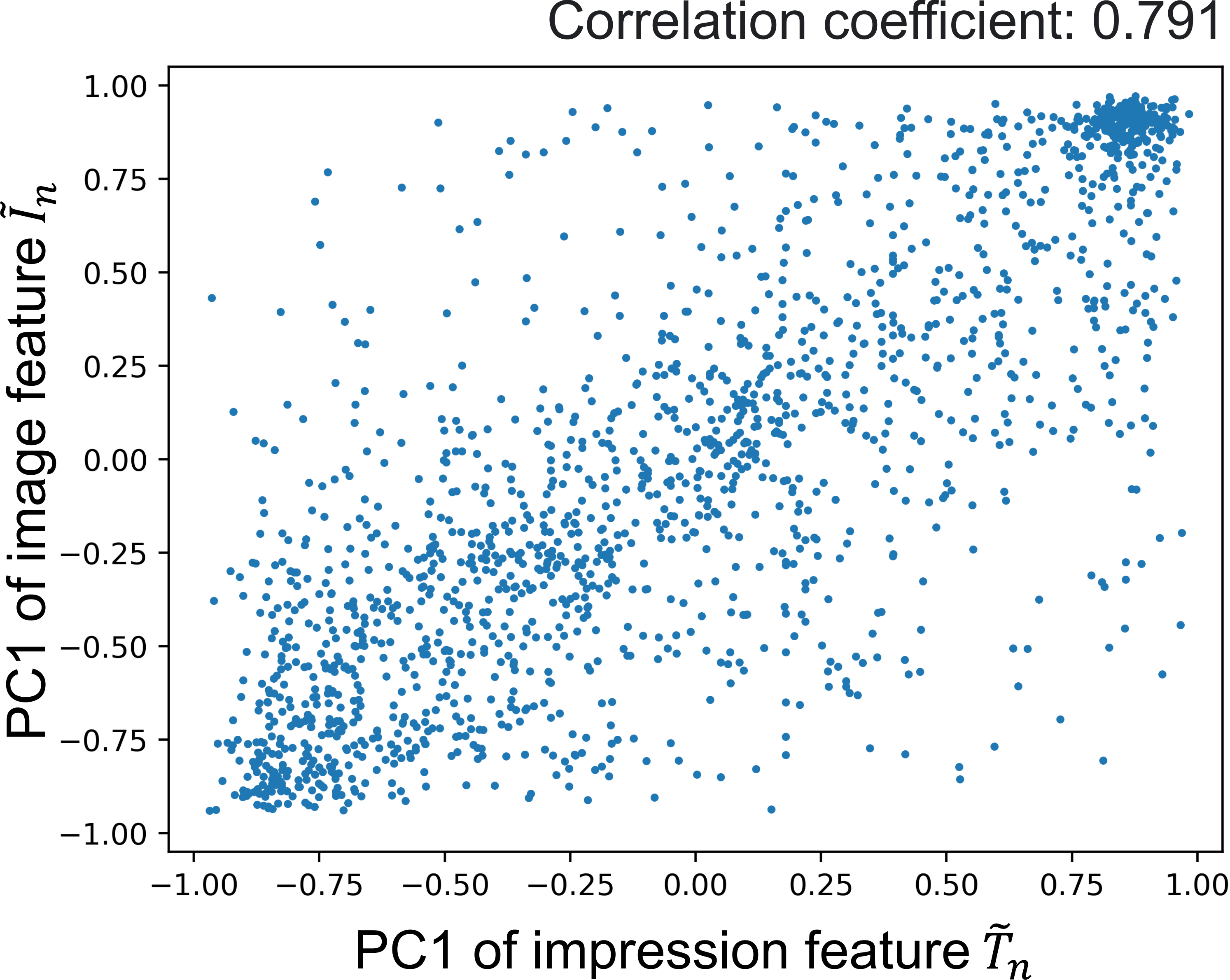}\\[-1mm]
            \subcaption{After contrastive learning}
            \label{imp-img_CLIP_after (Set)}
        \end{minipage} \\[6mm]
    \end{tabular}
    \caption{Visualization of feature distributions before and after contrastive learning by Impression-CLIP.}
    \label{fig:imp-img-plane}
\end{figure}

Fig.~\ref{fig:imp-img-plane} shows the feature distributions in the two-dimensional plane, where the horizontal and vertical axes show the first principal components (PC1s) of impression features ($T_n$ or $\tilde{T}_n$) and image features ($I_n$ or $\tilde{I}_n$), respectively. More specifically, Fig.~\ref{fig:imp-img-plane}(a) shows the distribution before contrastive learning; that is, the distribution of $(T_n, I_n)$, whereas (b) shows
the distribution after contrastive learning, that is $(\tilde{T}_n, \tilde{I}_n)$. 
The distributions clearly show that contrastive learning could enhance the cross-modal correlation between two modalities. The distribution of (a) is rather scattered, and no clear correlation between them. In contrast, (b) shows a near-diagonal distribution of $(\tilde{T}_n, \tilde{I}_n)$ and indicates $\tilde{T}_n \sim \tilde{I}_n$. This correlation is also confirmed quantitatively by the correlation coefficient of 0.791 in this two-dimensional representation.
\par

\subsection{Quantitative Evaluation}

Table~\ref{table:quantitative} shows the quantitative evaluation results on the MyFonts dataset, with the metrics described in Section~\ref{sec:metrics}. Impression-CLIP performs better in both directions (i.e., tag$\to$img and img$\to$tag) and evaluation metrics (ARR and mAP) than Cross-AE. This indicates that contrastive learning and the CLIP text encoder help to have better representation that satisfies $\tilde{I}_n\sim\tilde{T}_n$ for each font $F_n$ and $\tilde{I}_n\not\sim\tilde{T}_m$ for $F_n$ and $F_m$ ($n\neq m)$.
\par
In the evaluation by ARR. Impression-CLIP shows a symmetric behavior for images and impressions. In both directions, it equally achieves about 230 ARR. This fact indicates that the contrastive learning framework of CLIP helps to treat both modalities equally. In contrast, Cross-AE shows an asymmetric behavior where the image modality seems to have a higher priority. Note that mAP is an inherently asymmetric metric by the difference in the number of retrieval targets. (mAP (img$\to$tag) is easier than mAP (tag$\to$img) because the former chooses one from 233 tags and the latter from 1,710 fonts.)
\par
%
Impression-CLIP achieved not very high mAP, although it is still much better than Cross-AE. One possible reason for this performance limitation is the heavily imbalanced nature of the  
impression tags in the MyFonts dataset. Even after the elimination of less frequent tags ($<$\#50 occurrences), 70.0\% (38.2\%) tags occur less than 300 (100) times in the entire training dataset and thus are insufficient to train Impression-CLIP. In fact, if we plot a scattered plot between tag occurrences and average precision (AP), more frequent tags had higher APs.
\par

Table~\ref{table:quantitative} also shows the additional results on the AMT dataset.
As noted in Section~\ref{sec:dataset}, the AMT dataset contains 1,041 groups, and each group contains one font strongly correlated to a tag and two fonts weakly correlated to the tag. By using the impression tag, we seek the most similar font among the three fonts by a nearest-neighbor search. The search result is successful if the most correlated font is selected as the nearest neighbor and unsuccessful if one of two other weakly relevant fonts is selected. 
These two other weakly relevant fonts can occasionally have the impression tag as a noisy tag.
Therefore, better results indicate robustness to noise.
The accuracy $\in [0,1]$ in the table is the success rate. Impression-CLIP achieved 0.452, whereas Cross-AE did 0.383, which is almost the same as the chance rate (1/3=0.333). 
Our superiority is also confirmed by ARR $\in [1,3]$. 
\par

\begin{table}[t]
  \caption{Quantitative evaluation. ARR indicates average retrieval rank. ``tag$\to$img'' and ``img$\to$tag'' indicate a font image retrieval from impressions and vice versa.}
  \label{table:quantitative}
  \centering
  \begin{tabular}{l||>{\raggedleft}p{15mm}|>{\raggedleft}p{15mm}|>{\raggedleft}p{17mm}|>{\raggedleft}p{17mm}|r|r} \hline
  &\multicolumn{4}{c|}{MyFonts} &\multicolumn{2}{c}{AMT dataset} \\ 
          \cline{2-7}
          &\multicolumn{2}{c|}{ ARR$\downarrow$} &\multicolumn{2}{c|}{mAP$\uparrow$} &\multicolumn{1}{c|}{Accuracy$\uparrow$} &\multicolumn{1}{c}{ARR$\downarrow$}\\ 
          \cline{2-7}
    
           Method & tag$\to$img  & img$\to$tag & tag$\to$img & img$\to$tag &tag$\to$img & tag$\to$img\\
    \hline
    Impression-CLIP~   & $\mathbf{233.9}$   &  $\mathbf{232.9}$    &  $\mathbf{0.078}$    &   $\mathbf{0.170}$   &   $\mathbf{0.452}$       &  $\mathbf{1.802}$\\
    Cross-AE~\cite{kang2022shared}   &   618.1   &  357.9    &  0.054    &   0.133 &   0.383       &  1.937\\
    \hline
  \end{tabular}
\end{table}


\begin{figure}[t] 
\centering
\includegraphics[width=1\textwidth]{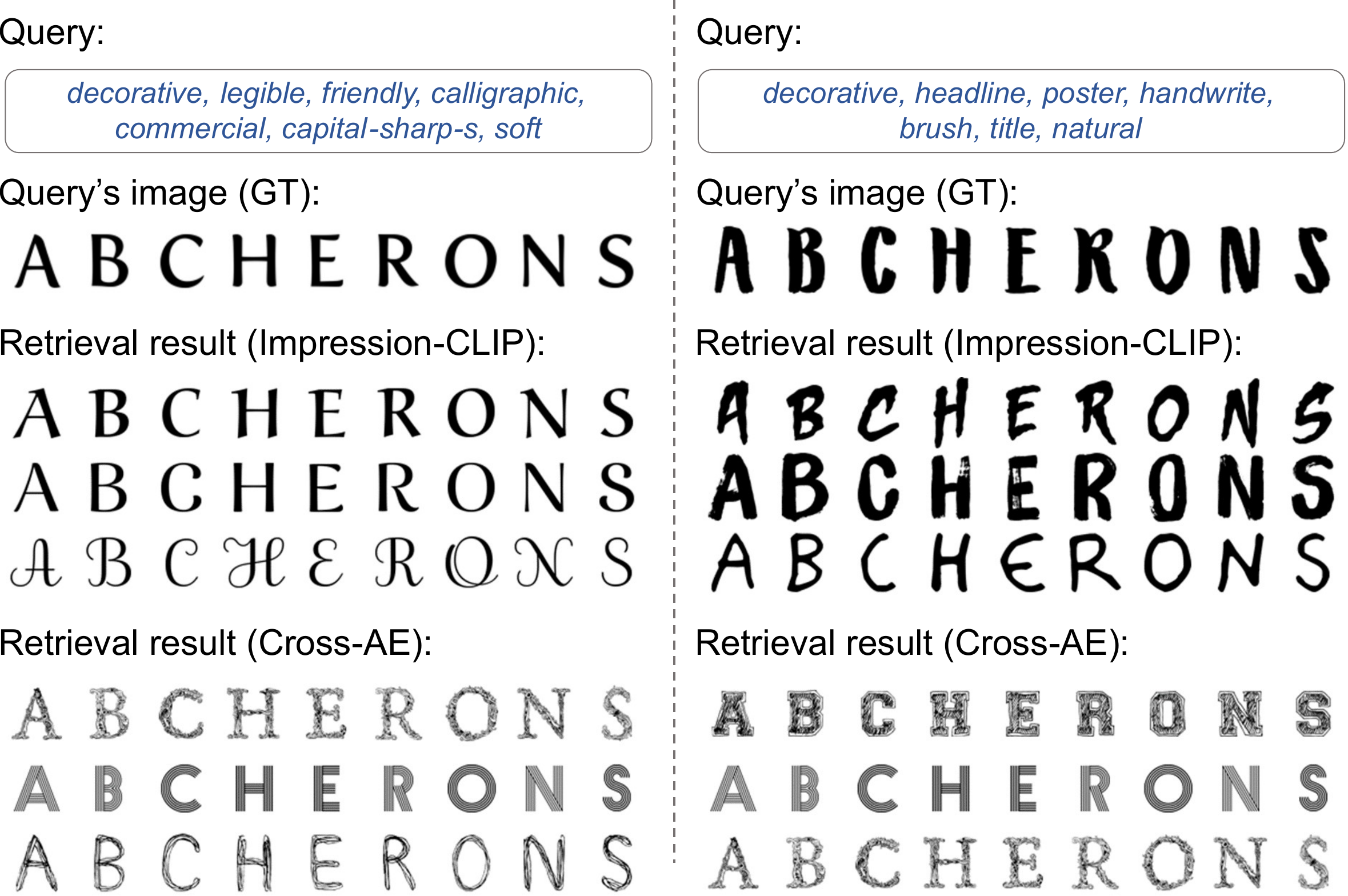}
\caption{Font image retrieval result by a set of impression tags. The three fonts retrieved are listed from top to bottom in order of top 1 to 3. Note the letters ``ABC'' and ``HERONS'' are selected to observe various local shape variations, such as curves and corners.} \label{fig:imp2img}
\end{figure} 

\subsection{Qualitative Evaluation}

Fig.~\ref{fig:imp2img} shows the font image retrieval results within the 1710-font test set. Specifically, we pick up one font $F_n$ and derive its impression vector $\tilde{T}_n$. Then, we find three fonts whose $\tilde{I}_m$ has top-3 similarity with $\tilde{T}_n$.
At least, the top and the second-top fonts selected by Impression-CLIP are similar to the ground truth (GT) font where the query impression vector $\tilde{T}_n$ is derived. 
In contrast, Cross-AE could not retrieve fonts similar to GT.
\par

\begin{figure}[t] 
\centering
\includegraphics[width=\textwidth]{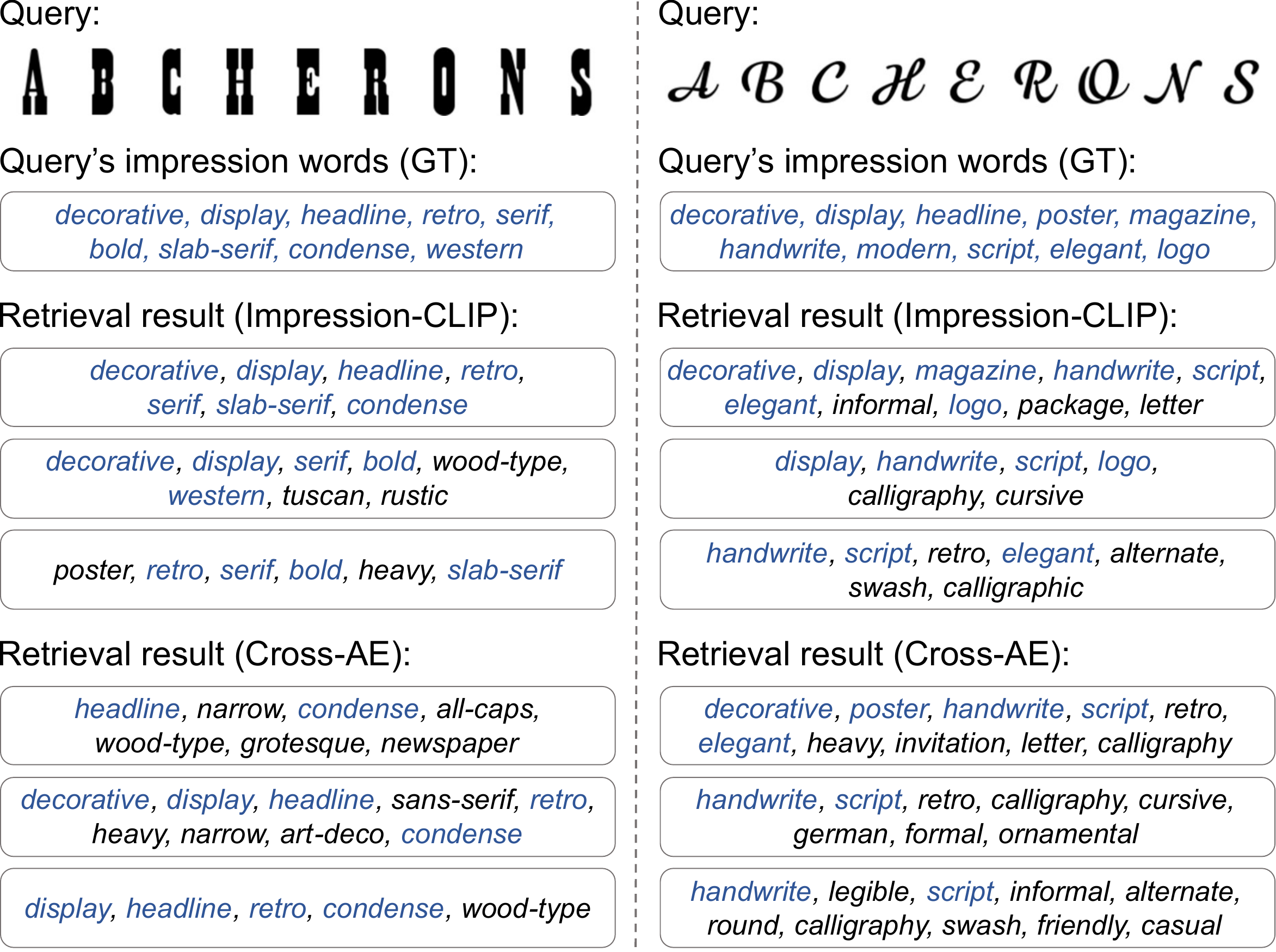}
\caption{Impression retrieval (i.e., estimation) result by a query font image. The three impression sets retrieved are listed from top to bottom in order of top 1 to 3.} \label{fig:img2imp}
\end{figure} 
Fig.~\ref{fig:img2imp} shows the impressions estimated by a font image feature $\tilde{I}_n$. These results show the tag sets of fonts, each of whose impression feature $\tilde{T}_k$ has top-3 similarity to $\tilde{I}_n$.
Like the image retrieval result, Impression-CLIP successfully estimates more 
correct impressions than Cross-AE. Even though the estimated results show spurious tags, such as \Tag{heavy}, they still seem reasonable because of the existence of similar impressions (\Tag{bold}) in GT. In contrast, Cross-AE retrieved the incoherent impressions or even did not fit the query image. For example, in the left example, the results by Cross-AE include
\Tag{sans-serif} tag, whereas this query image has heavy serifs. 

\begin{figure}[t] 
\centering
\includegraphics[width=\textwidth]{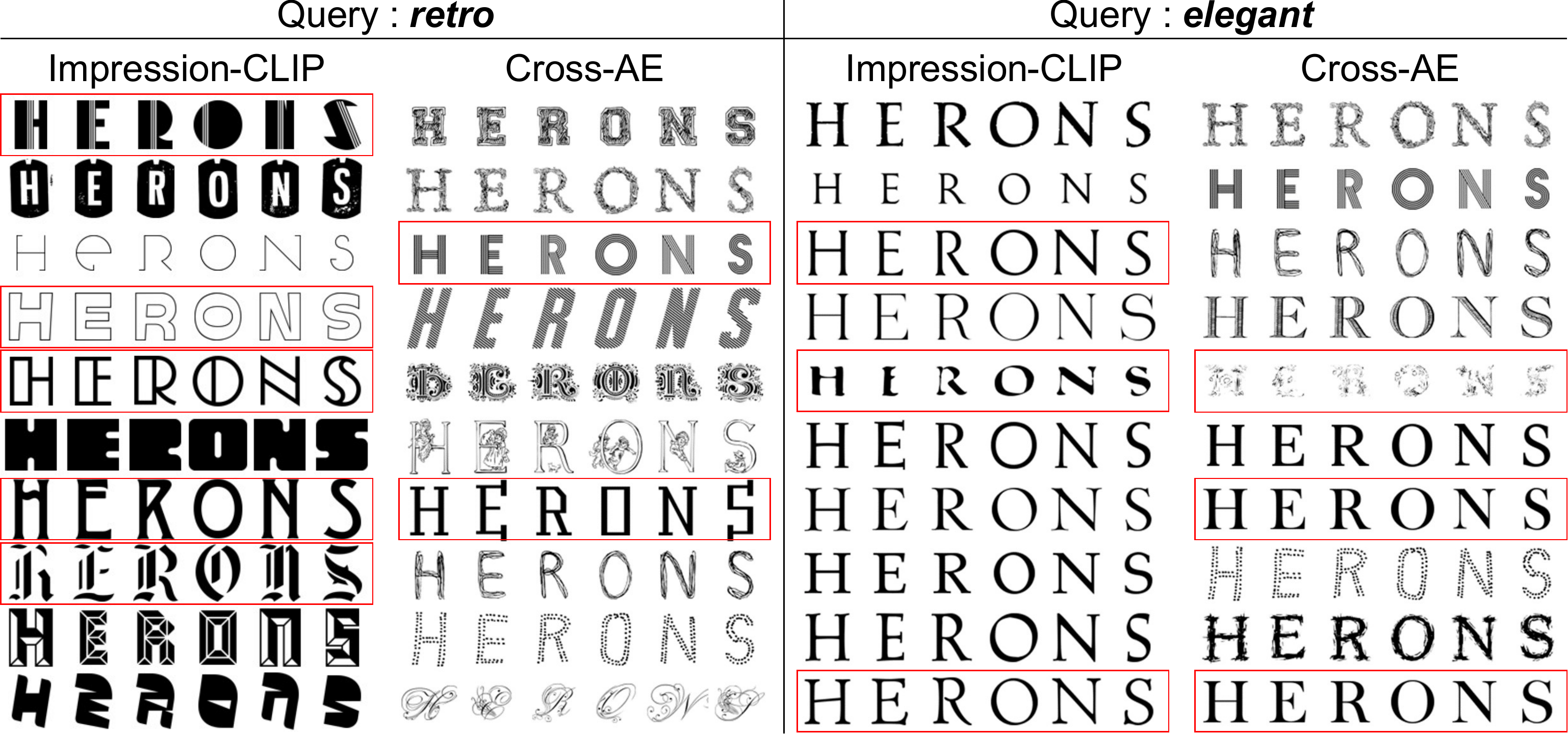}
\caption{Font image retrieval result by a tag. The red boxes are
samples that contained the query tag.} \label{fig:single_tag2img}
\end{figure} 


Fig.~\ref{fig:single_tag2img} shows the retrieved font images by a single impression tag. Here, \Tag{retro} and \Tag{elegant} are used, respectively.
In the example of \Tag{retro}, Within 10 retrieved fonts, Impression-CLIP successfully retrieves 6 fonts with the \Tag{retro} impression, whereas Cross-AE retrieves only 2 fonts. Moreover, the retrieved results by Impression-CLIP show diverse styles. On the other hand, the results for  \Tag{elegant} show that both models retrieve only three ``correct'' fonts among their 10 candidates. However, the candidates by Impression-CLIP are often similar to each other. Consequently, it seems that Impression-CLIP could retrieve fonts that miss the correct tag \Tag{elegant}.
\par




\begin{figure}[t] 
    \includegraphics[width=\textwidth]{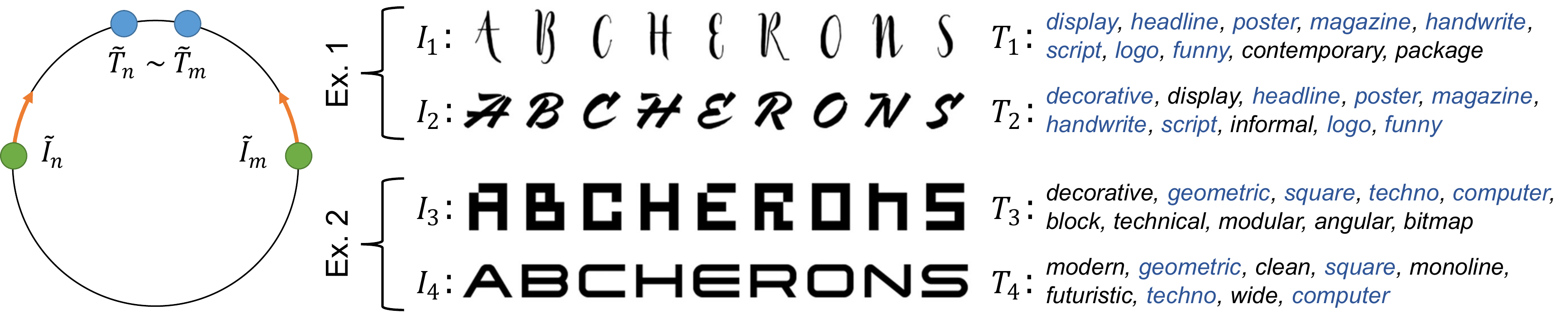}
    \caption{Font images that become closer by their impressions. 
    } \label{fig:approached_images}

\end{figure} 

\begin{figure}[t] 
\includegraphics[width=\textwidth]{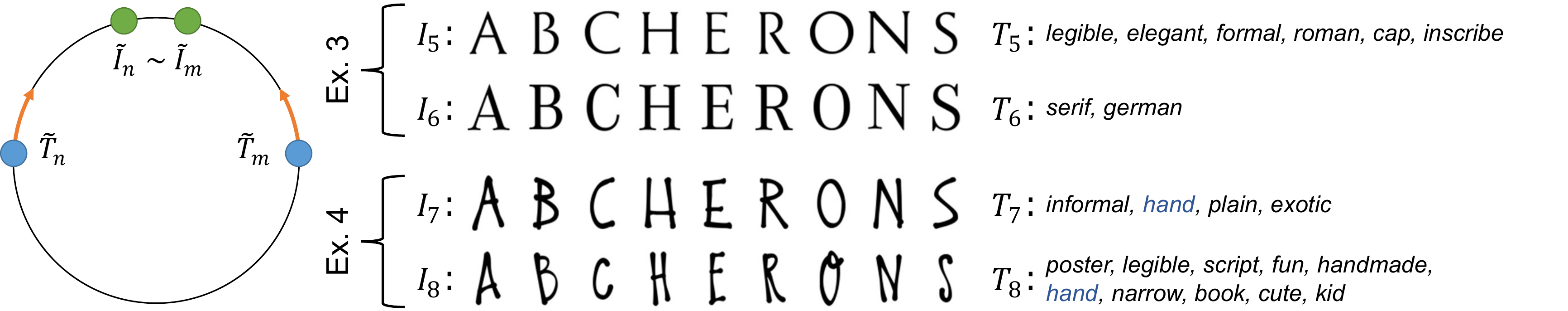}
\caption{Font impressions that become closer by their images. 
} \label{fig:approached_impressions}
\end{figure} 

\begin{figure}[t]
    \begin{tabular}{cc}
        \begin{minipage}[t]{0.47\textwidth}
            \centering
            \includegraphics[width=\linewidth]{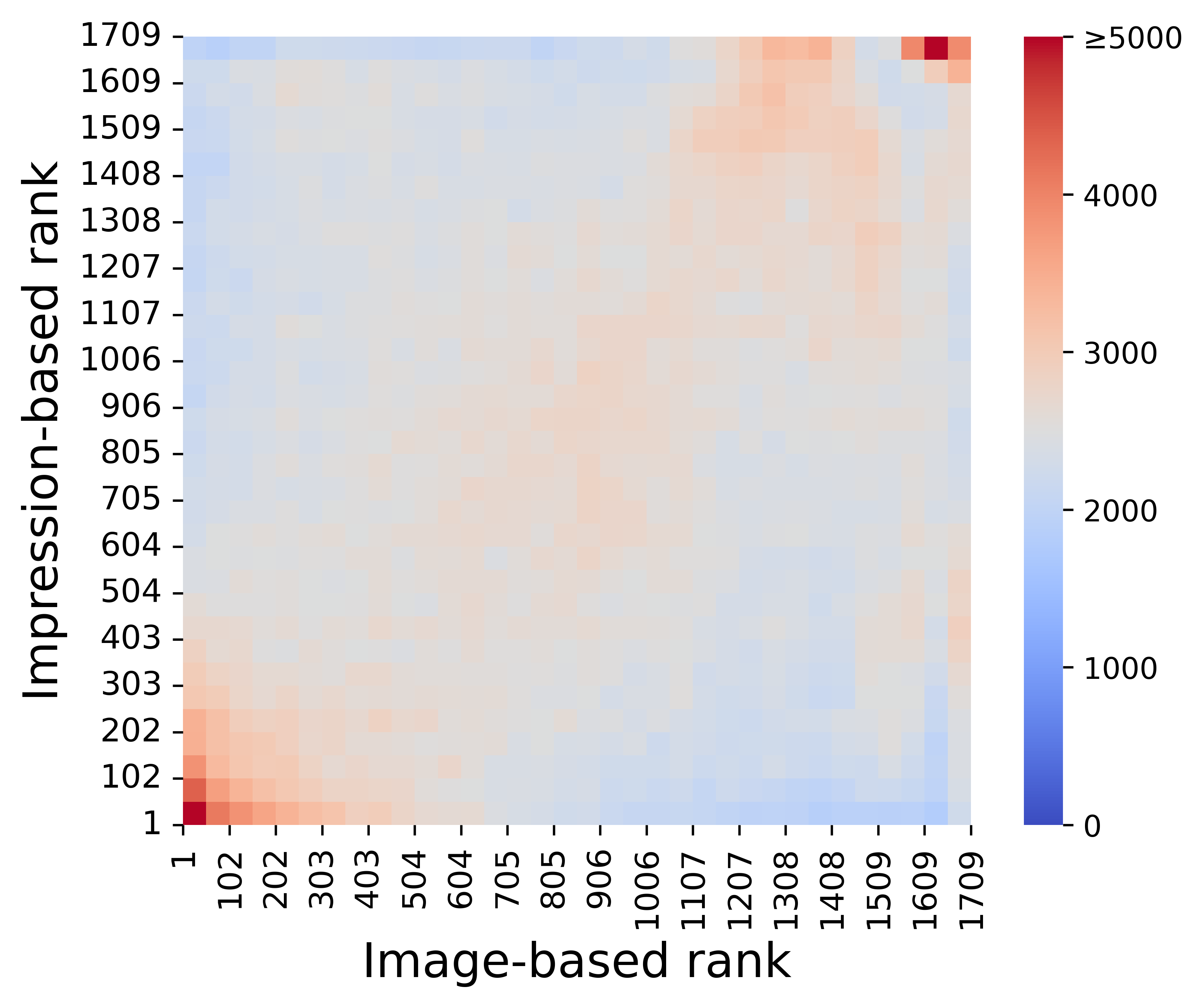} \\
            \subcaption{Before co-embedding} 
            \label{fig:rank_matrix_before}
        \end{minipage} &
        
        \hspace{0.035\columnwidth}
        
        \begin{minipage}[t]{0.47\textwidth}
            \centering
            \includegraphics[width=\linewidth]{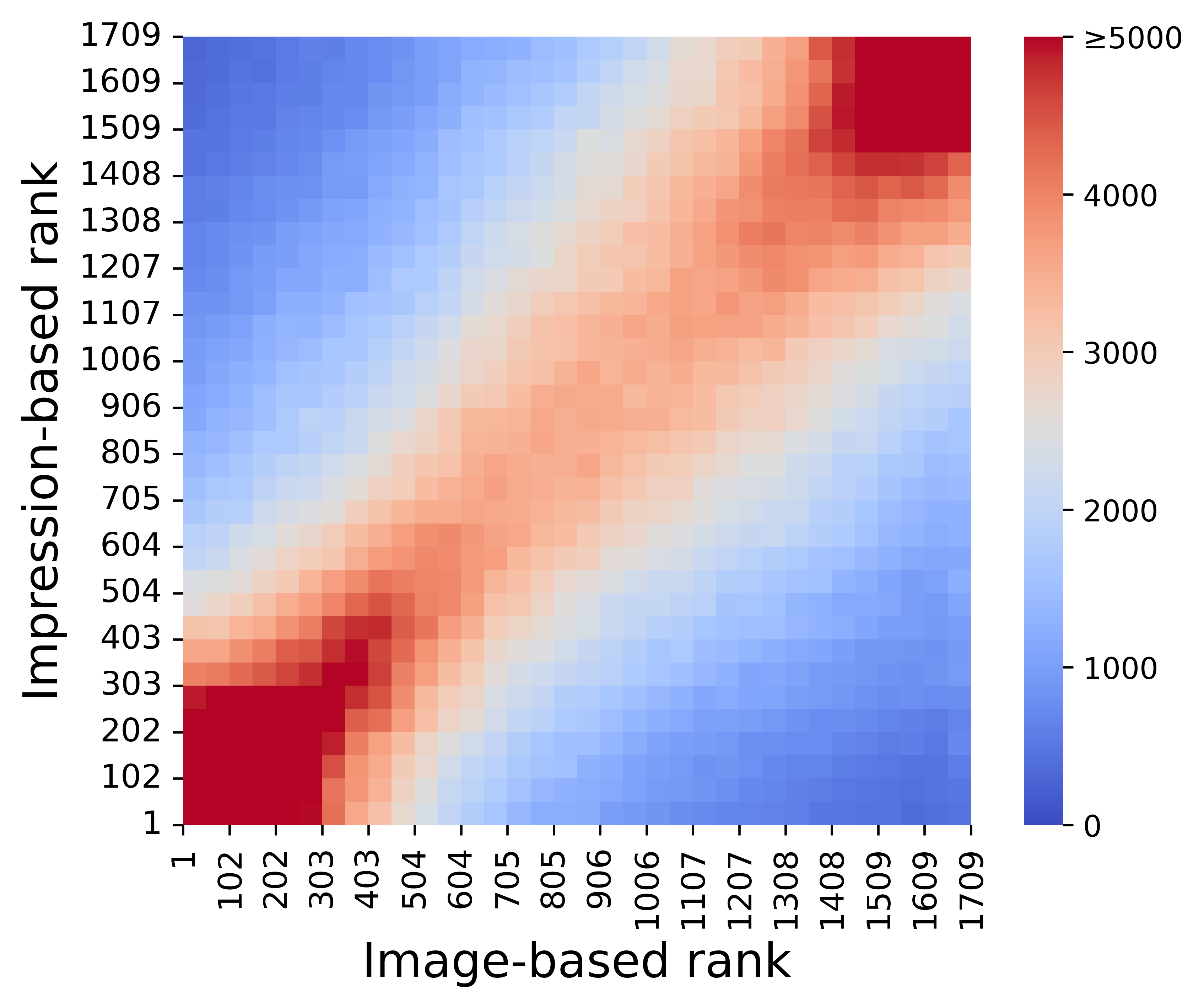}\\
            \subcaption{After co-embedding}
            \label{fig:rank_matrix_after}
        \end{minipage}
    \end{tabular}
    \caption{Histogram of rank-pairs. See the main text for the details.}
    \label{fig:rank_matrix}
\end{figure}

\subsection{Consistency of Similarities in Two Modalities}  

Impression-CLIP performs cross-modal co-embedding with contrastive learning and, therefore, shows synergistic effects between the two modalities. Fig.~\ref{fig:approached_images} shows two examples where a pair of very different font images ($I_n\not\sim I_m$) are embedded closer (i.e., pulled together) by similar impressions. Formally,  $I_n \not\sim I_m$ but $\tilde{I}_n \sim \tilde{I}_m$ because $\tilde{T}_n \sim \tilde{T}_m$. More simply, if we only observe these two font images, they are totally different; however, with the help of the impression modality, these font images newly get a serendipitous relationship, $\tilde{I}_n \sim \tilde{I}_m$.
\par

Similarly, we can also find serendipitous relationships between impressions. 
Fig.~\ref{fig:approached_impressions} shows two examples where a pair similar images $\tilde{I}_n \sim \tilde{I}_m$ pull together two very different impressions ($T_n\not\sim T_m$) to be $\tilde{T}_n \sim \tilde{T}_m$. In fact, the above example has no overlap between their tags \{\Tag{legible}, \Tag{elegant}, \ldots, \Tag{inscribe}\} and \{\Tag{serif}, \Tag{german}\}
although their images are similar ($I_n\sim I_m$). As a result of cross-modal contrastive learning, their impressions could become closer and we can know the upper font should have \Tag{serif} like the lower font. \par
%
As we observed in Figs.~\ref{fig:approached_images} and \ref{fig:approached_impressions}, the similarity in one modality changes by the similarity in the other modality, and we can quantify this change in terms of {\em similarity ranking}. Here is an example using the upper pair of Fig.~\ref{fig:approached_images}.
\begin{itemize}
    \item About the images, $I_2$ is ranked as the 1,624th image for $I_1$ among 1,709 test fonts; namely, $I_2$ is very different for $I_1$. However, after co-embedding, $\tilde{I}_2$ becomes the 14th image for $\tilde{I}_1$. 
    \item About the impressions, $T_2$ is ranked as the 1st impression (i.e., the most similar impression) for $T_1$; after co-embedding, $\tilde{T}_2$ becomes the 2nd impression for $\tilde{T}_1$. 
\end{itemize}
Here is another example using the upper pair of Fig.~\ref{fig:approached_impressions}.
\begin{itemize}
\item $I_6$ is ranked as the 49th image for $I_5$, 
and $\tilde{I}_6$ becomes 18th image for $\tilde{I}_5$.
\item $T_6$ is ranked as the 1,523th impression for $T_5$. However, by the image similarity, $\tilde{T}_6$ becomes the 14th impression for $\tilde{T}_5$.
\end{itemize}
\par
%
We now describe the ranking change by co-embedding using {\em rank-pairs} as $(1624, 1)\to (14, 2)$ for the first example. The former value in each rank-pair is the image-based rank, and the latter is the impression-based. Following the same way, the second example is described as $(49, 1523)\to (18, 14)$. In these examples, there are large differences in the rank-pairs before co-embedding, such that $(1624, 1)$ and $(49, 1523)$. However, after co-embedding, the differences are minimized as $(14, 2)$ and $(18, 14)$. This means that similarity becomes more consistent over modalities by the effect of cross-modal co-embedding by Impression-CLIP.
\par

Fig.~\ref{fig:rank_matrix} shows histograms of rank-pairs before and after co-embedding. 
Before co-embedding, the histogram shows scattered distributions. This means that the similarities between fonts are different (i.e., inconsistent) between the image and impression modality. In contrast, after co-embedding, the histogram shows diagonal peaks. This means that the font similarities become much more consistent between the two modalities. Simply saying, ``If a pair of fonts has high image similarity, their impression similarity is also high.''  Fig.~\ref{fig:rank_matrix} proves that our Impression-CLIP realizes such a consistent cross-modal co-embedding of fonts.

\section{Conclusion, Limitation, and Future Work}
We proposed Impression-CLIP, which is a novel multimodal neural network based on CLIP, to associate font shapes and their impressions through co-embedding. 
We showed that Impression-CLIP could capture the unstable and weak correlations between font shapes and impressions, utilizing the merits of contrastive learning. Further, through quantitative and qualitative evaluations of cross-modal retrieval, we also confirmed that not only Impression-CLIP has better performance than the state-of-the-art co-embedding model called Cross-AE, but it also has robustness to noise and missing tags. 
\par

The current limitation is the imbalance issue of the impression tags. There are many minor tags that cause degradation in the retrieval experiments. We will introduce some tricks to balance tag occurrences while also caring for noisy and missing tags. 


\par
\bigskip
\noindent{\bf Acknowledgment}:\ This work was supported by JSPS KAKENHI Grant Number JP22H00540.



%
%
%





\bibliographystyle{splncs04}
\bibliography{ref}

\begin{thebibliography}{10}
\providecommand{\url}[1]{\texttt{#1}}
\providecommand{\urlprefix}{URL }
\providecommand{\doi}[1]{https://doi.org/#1}

\bibitem{chen2019large}
Chen, T., Wang, Z., Xu, N., Jin, H., Luo, J.: Large-scale tag-based font retrieval with generative feature learning. In: Proceedings of the IEEE/CVF International Conference on Computer Vision (ICCV). pp. 9116--9125 (2019)

\bibitem{choi2019emotype}
Choi, S., Aizawa, K.: Emotype: Expressing emotions by changing typeface in mobile messenger texting. Multimedia Tools and Applications  \textbf{78}(11),  14155--14172 (2019)

\bibitem{choi2019assist}
Choi, S., Matsumura, S., Aizawa, K.: Assist users' interactions in font search with unexpected but useful concepts generated by multimodal learning. In: Proceedings of the International Conference on Multimedia Retrieval (ICMR). pp. 235--243 (2019)

\bibitem{Conde_2021_CVPR}
Conde, M.V., Turgutlu, K.: Clip-art: Contrastive pre-training for fine-grained art classification. In: Proceedings of the IEEE/CVF Conference on Computer Vision and Pattern Recognition Workshops (CVPRW). pp. 3956--3960 (2021)

\bibitem{davis1933determinants}
Davis, R.C., Smith, H.J.: Determinants of feeling tone in type faces. Journal of Applied Psychology  \textbf{17}(6), ~742 (1933)

\bibitem{poffenberger1923study}
Franken, R.: A study of the appropriateness of type faces. Journal of Applied Psychology  \textbf{7}(4), ~312 (1923)

\bibitem{frans2022clipdraw}
Frans, K., Soros, L., Witkowski, O.: Clipdraw: Exploring text-to-drawing synthesis through language-image encoders. In: Advances in Neural Information Processing Systems (NeurIPS). pp. 5207--5218 (2022)

\bibitem{henderson2004impression}
Henderson, P.W., Giese, J.L., Cote, J.A.: Impression management using typeface design. Journal of Marketing  \textbf{68}(4),  60--72 (2004)

\bibitem{izumi2022zero}
Izumi, K., Yanai, K.: Zero-shot font style transfer with a differentiable renderer. In: Proceedings of the 4th ACM International Conference on Multimedia Asia (MM Asia). pp.~1--5 (2022)

\bibitem{kang2022shared}
Kang, J., Haraguchi, D., Matsuda, S., Kimura, A., Uchida, S.: Shared latent space of font shapes and their noisy impressions. In: Proceedings of the 28th International Conference on Multimedia Modeling (MMM). pp. 146--157 (2022)

\bibitem{kulahcioglu2020fonts}
Kulahcioglu, T., De~Melo, G.: Fonts like this but happier: A new way to discover fonts. In: Proceedings of the 28th ACM International Conference on Multimedia (MM). pp. 2973--2981 (2020)

\bibitem{Kwon_2022_CVPR}
Kwon, G., Ye, J.C.: Clipstyler: Image style transfer with a single text condition. In: Proceedings of the IEEE/CVF Conference on Computer Vision and Pattern Recognition (CVPR). pp. 18062--18071 (2022)

\bibitem{luo2022clip4clip}
Luo, H., Ji, L., Zhong, M., Chen, Y., Lei, W., Duan, N., Li, T.: Clip4clip: An empirical study of clip for end to end video clip retrieval and captioning. Neurocomputing  \textbf{508},  293--304 (2022)

\bibitem{ma2022x}
Ma, Y., Xu, G., Sun, X., Yan, M., Zhang, J., Ji, R.: X-clip: End-to-end multi-grained contrastive learning for video-text retrieval. In: Proceedings of the 30th ACM International Conference on Multimedia (MM). pp. 638--647 (2022)

\bibitem{matsuda2021impressions2font}
Matsuda, S., Kimura, A., Uchida, S.: Impressions2font: Generating fonts by specifying impressions. In: Proceedings of the 16th International Conference on Document Analysis and Recognition (ICDAR). pp. 739--754 (2021)

\bibitem{matsuda2022font}
Matsuda, S., Kimura, A., Uchida, S.: Font generation with missing impression labels. In: Proceedings of the 26th International Conference on Pattern Recognition (ICPR). pp. 1400--1406 (2022)

\bibitem{o2014exploratory}
O'Donovan, P., L{\=\i}beks, J., Agarwala, A., Hertzmann, A.: Exploratory font selection using crowdsourced attributes. ACM Transactions on Graphics (TOG)  \textbf{33}(4), ~1--9 (2014)

\bibitem{radford2021learning}
Radford, A., Kim, J.W., Hallacy, C., Ramesh, A., Goh, G., Agarwal, S., Sastry, G., Askell, A., Mishkin, P., Clark, J., et~al.: Learning transferable visual models from natural language supervision. In: Proceedings of the 38th International Conference on Machine Learning (ICML). pp. 8748--8763 (2021)

\bibitem{Rombach_2022_CVPR}
Rombach, R., Blattmann, A., Lorenz, D., Esser, P., Ommer, B.: High-resolution image synthesis with latent diffusion models. In: Proceedings of the IEEE/CVF Conference on Computer Vision and Pattern Recognition (CVPR). pp. 10684--10695 (2022)

\bibitem{ueda2021parts}
Ueda, M., Kimura, A., Uchida, S.: Which parts determine the impression of the font? In: Proceedings of the 16th International Conference on Document Analysis and Recognition (ICDAR). pp. 723--738 (2021)

\bibitem{ueda2022font}
Ueda, M., Kimura, A., Uchida, S.: Font shape-to-impression translation. In: Proceedings of 15th IAPR International Workshop on Document Analysis Systems (DAS). pp. 3--17 (2022)

\bibitem{wang2020attribute2font}
Wang, Y., Gao, Y., Lian, Z.: Attribute2font: Creating fonts you want from attributes. ACM Transactions on Graphics (TOG)  \textbf{39}(4),  1--15 (2020)

\end{thebibliography}
\end{document}